\RequirePackage[l2tabu,orthodox]{nag}

\documentclass[
  paper       = a4,
  headheight  = 16pt,
  footheight  = 16pt,
  fontsize    = 10pt,
  twoside     = true,
  titlepage   = true,
]{scrartcl}

\usepackage{changepage}
\usepackage{float}
\usepackage{fontspec}
\usepackage[ngerman,british]{babel}
\usepackage[english=british]{csquotes}
\setquotestyle[guillemets]{german}

\usepackage[
  backend=biber,
  natbib=true,
  citestyle=authoryear,
  style=authoryear,
  maxcitenames=1,
  maxbibnames=5,
  uniquename=init,
  giveninits=true,
  sortlocale={de_DE_phonebook},
]{biblatex}

\DefineBibliographyExtras{british}{%
}

\DeclareFieldFormat{eprint:urn}{%
  \textsc{\scriptsize{URN\addcolon}}\space
  \ifhyperref
    {\href{http://www.nbn-resolving.org/#1}{\nolinkurl{#1}}}
    {\nolinkurl{#1}}
}

\DeclareFieldFormat{eprint:hdl}{%
  \textsc{\scriptsize{HDL\addcolon}}\space
  \ifhyperref
    {\href{http://hdl.handle.net/#1}{\nolinkurl{#1}}}
    {\nolinkurl{#1}}
}

\bibsetup{
  \setcounter{abbrvpenalty}{10000}
  \setcounter{highnamepenalty}{10000}
  \setcounter{lownamepenalty}{10000}
}

\AtEveryBibitem{\clearfield{month}}
\AtEveryBibitem{\clearfield{day}}

\usepackage[%
  paperwidth  = 17cm,
  paperheight = 24cm,
  top         = 25mm,
  bottom      = 24mm,
  outer       = 15mm,
  inner       = 23mm,
  footskip    = 10mm,
]{geometry}
\newlength{\secnumwidth}
\newlength{\subsecnumwidth}
\newlength{\subsubsecnumwidth}

\usepackage{xcolor}

\usepackage{setspace}
\deffootnote[1em]{1em}{0pt}{\textsuperscript{\thefootnotemark}\ }

\usepackage{graphicx}
\usepackage{subfig}

\usepackage{placeins}

\usepackage{mathtools}
\usepackage{amsfonts}
\usepackage{amssymb}

\usepackage[font={footnotesize}]{caption}
\setcapwidth{0.8\textwidth}
\setcapindent{0pt}
\addtokomafont{caption}{\centering}
\addtokomafont{captionlabel}{\bfseries}

\colorlet{commentcolour}{green!50!black}
\colorlet{stringcolour}{red!60!black}
\colorlet{keywordcolour}{magenta!90!black}
\colorlet{exceptioncolour}{yellow!50!red}
\colorlet{commandcolour}{blue!60!black}
\colorlet{numpycolour}{blue!60!green}
\colorlet{literatecolour}{magenta!90!black}
\colorlet{promptcolour}{green!50!black}
\colorlet{specmethodcolour}{violet}

\usepackage{textcomp} 
\usepackage{listings}
\makeatletter
\lstdefinestyle{mylststyle}{
  basicstyle=%
    \ttfamily
    \lst@ifdisplaystyle\footnotesize\fi
}
\makeatother
\usepackage[%
  colorlinks  = true,%
  allcolors   = black,%
  urlcolor    = black,%
  pdfencoding = auto,%
  hyperindex,%
  pageanchor,%
  pdfdisplaydoctitle,%
  psdextra,%
  frenchlinks=true,
  final%
]{hyperref}

\usepackage{doi}

\usepackage[english,nameinlink]{cleveref}
\crefname{figure}{Figure}{Figures}
\Crefname{figure}{Figure}{Figures}
\crefname{table}{Table}{Tables}
\Crefname{table}{Table}{Tables}
\crefname{section}{Section}{Sections}
\Crefname{section}{Section}{Sections}
\crefname{subsection}{Subsection}{Subsections}
\Crefname{subsection}{Subsection}{Subsections}
\crefname{listing}{Listing}{Listings}
\Crefname{listing}{Listing}{Listings}

\usepackage[headsepline]{scrlayer-scrpage}
\clearpairofpagestyles
\setkomafont{pageheadfoot}{\footnotesize \rmfamily}
\setkomafont{pagination}{\footnotesize \rmfamily}
\setkomafont{title}{\linespread{1} \huge \rmfamily \textbf}
\setkomafont{subtitle}{\linespread{1} \large \rmfamily \textbf}
\setkomafont{author}{{\linespread{1} \rmfamily}}
\setkomafont{date}{\rmfamily}
\setkomafont{dedication}{{\linespread{1} \rmfamily}}
\setkomafont{publishers}{\rmfamily}
\setkomafont{subject}{\rmfamily}
\setkomafont{titlehead}{\rmfamily}

\DeclareNewLayer[%
  headsep,
  foreground,
  contents={%
    \begin{minipage}[t]{\textwidth}%
      \usekomafont{titlehead}{\csname @titlehead\endcsname\par}%
    \end{minipage}\par}
]{titlehead}
\DeclarePageStyleByLayers{titlepage}{titlehead}
\AddToHook{cmd/maketitle/after}{\thispagestyle{titlepage}}

\makeatletter
\def\@listI{\leftmargin\leftmargin
            \parsep 1\p@ \@plus2\p@ \@minus\p@
            \topsep 8\p@ \@plus2\p@ \@minus4\p@
            \itemsep1\p@ \@plus2\p@ \@minus\p@}
\makeatother

\newcommand*{\mainauthor}[1]{\def\MainAuthor{#1}}

\newcommand*{\headtitle}[1]{\def\HeadTitle{#1}}
\newcommand*{\maintitle}[1]{\def\MainTitle{#1}}

\newcommand*{\fulltitle}[1]{\def\FullTitle{#1}}

\newcommand*{\orcid}[1]{\href{#1}{\includegraphics[height=10pt]{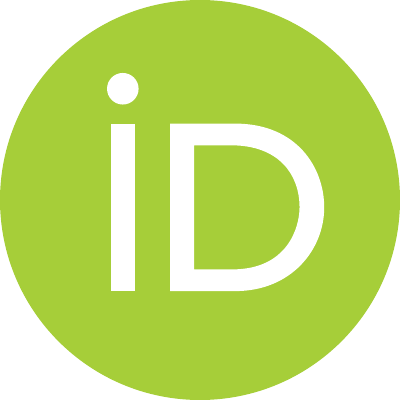}\,\nolinkurl{#1}}}
\newcommand*{\orcidlogo}[1]{\href{#1}{\includegraphics[height=10pt]{orcidlogo.pdf}}}

\newcommand*{\email}[1]{\href{mailto:#1}{\nolinkurl{#1}}}

\newcommand*{\pagestart}[1]{\def\Page{#1}}
\newcommand*{\keywords}[1]{\def\Keywords{#1}}

\makeatletter
\renewcommand{\@pnumwidth}{3em}
\renewcommand{\@tocrmarg}{4em}
\makeatother

\ofoot[]{\pagemark}

\date{}

\hypersetup{
  pdfsubject  = {Proceedings of the 10th bwHPC Symposium, Freiburg September 2024},
  pdfkeywords = {bwHPC, HPC, 10th bwHPC Symposium},
}

\mainauthor{Frank Mollard}
\maintitle{Adversarial Evasion Attacks on Computer Vision using SHAP Values}
\fulltitle{\MainTitle}
\headtitle{\MainTitle}
\pagestart{51}

\usepackage{amsmath,amsthm,setspace}

\DeclareMathOperator\sign{sign}

\begin{document}

\begin{titlepage}
\thispagestyle{myheadings}
\raggedright
\vspace*{.05\textheight}

{\usekomafont{title}
\MainTitle 
\par}


\vspace{\baselineskip}


 \vspace{.5\baselineskip}

{\usekomafont{author}
\hyperref[sec:cauthor]{\MainAuthor}\textsuperscript{*}%
    \,\orcidlogo{https://orcid.org/0000-0002-6469-4764},
Marcus Becker\textsuperscript{\ddag}%
    \,\orcidlogo{https://orcid.org/0000-0001-5801-3785},
Florian Röhrbein\textsuperscript{\dag}%
    \,\orcidlogo{https://orcid.org/0000-0002-4709-2673}
\par}

\vspace{.5\baselineskip}
{\usekomafont{author}
\small
\textsuperscript{*}Business Intelligence \& Data Science, Infraserv GmbH \& Co. Höchst KG, Germany
\\
\small
\textsuperscript{\ddag}International School of Management, Germany
\\
\small
\textsuperscript{\dag}TU-Chemnitz, Germany
\par}

\vspace{\baselineskip}
\let\endtitlepage\relax
\end{titlepage}



\rohead{\HeadTitle} 
\lehead{Proceedings of the 10th bwHPC Symposium}

\hypersetup{
  pdftitle={\FullTitle},
  pdfauthor={\MainAuthor~et~al.},
  pdfkeywords={\Keywords},
}

\setcounter{page}{\Page}



\section*{Abstract}
The paper introduces a white-box attack on computer vision models using SHAP values. It demonstrates how adversarial evasion attacks can compromise the performance of deep learning models by reducing output confidence or inducing misclassifications. Such attacks are particularly insidious as they can deceive the perception of an algorithm while eluding human perception due to their imperceptibility to the human eye. The proposed attack leverages SHAP values to quantify the significance of individual inputs to the output at the inference stage. A comparison is drawn between the SHAP attack and the well-known Fast Gradient Sign Method. We find evidence that SHAP attacks are more robust in generating misclassifications particularly in gradient hiding scenarios.

\section{Introduction}\label{sec:intro}

Adversarial attacks on deep learning models, such as Artificial Neural Networks (ANN), aim to undermine performance by reducing output confidence or inducing misclassifications \citep{ZLC2022}. These attacks pose a serious threat because they deceive the perception of the algorithm while being deliberately designed to avoid detection by human perception. There are various access paths for an attack, with either the training data (poisoning), the model itself, or the application data (evasion) being targeted. When attacking the training data, the attacker must have access to it. This involves either changing or adding data. Evasion attacks, on the other hand, refer to the input of the trained model in the inference phase. The input is only manipulated slightly by adding imperceptible perturbations so the viewer will barely notice any differences. This makes evasion attacks very attractive for attackers, as they can remain unnoticed. We make a distinction between white-box and black-box attacks. White-box attacks require knowledge of the model and access to it \citep{HiAtPe2017, TrEt2016}. Black-box attacks, on the other hand, do not require any knowledge of the model. However, it is important that the input and output of the model must at least be observable \citep{PMG2017, papernot2016transferabilitymachinelearningphenomena, RoEt2017, TrEt2016, FrJhRi2015, ShEt2017, MoFaFr2016}. 

In the field of white-box attacks, there are several approaches, some of which differ significantly, reflecting the diversity of strategies explored in scientific research. One of the first papers dealing with white-box evasion attacks is by \textcite{szegedy2014intriguingpropertiesneuralnetworks}, in which the authors ensured that the model would misclassify a so-called adversarial example by optimising its distance from the correctly classified original input. \textcite{MoFaFr2016} proposes a method called DeepFool. \textcite{PapNicMcPaD2016} developed an algorithm that only changed around 4 \% of the input data to lead to miscalculation. \textcite{alaifari2019adefiterativealgorithmconstruct} added slight deformations to the input to keep the manipulation invisible, thereby inducing misclassifications. Probably the best-known method is called the Fast Gradient Sign Method (FGSM) by \textcite{goodfellow2015explainingharnessingadversarialexamples}. The advantage of this method lies in its relative simplicity of implementation compared to the previously discussed approaches. It leverages the fact that the optimisation relies on gradient descent. For every classification with a non-perfect result, there will be a gradient that is not equal to zero. This allows the sign of the gradient to be added to any observation $x$, reversing the gradient descent. 
\vskip -0.1in
\begin{equation}
x' = x + \varepsilon \cdot \sign(\nabla_x L(x, t, \omega))
\end{equation}
\vskip -0.15in
The observations $x$, the target $t$ and the model parameters $\omega$ are constant. As the gradient $\nabla$ of the loss function $L$ is close to zero in a good model, only the signs are used. The scalar $\varepsilon$ represents the magnitude of the attack. For instance, if the slope of the loss function is negative, $\varepsilon$ is subtracted from the pixel value $x$, vice versa. This causes a shift in the direction of increasing errors, resembling a reverse gradient descent process. Though, the $\varepsilon$ is assumed to be equal for all weights, even though the gradients may be different, potentially leading to overly strong or weak adjustments for the non-linear loss function. 
 
The FGSM prioritizes speed over efficiency \citep{carlini2017evaluatingrobustnessneuralnetworks}, which means that attacks often fail. \textcite{kurakin2017adversarialexamplesphysicalworld} enhance this method by incorporating an iterative approach, where the overall $\varepsilon$ is broken down into smaller increments, and the gradient is recalculated at each step. This accounts for the non-linearity of the loss functions of ANN. The modification substantially boosts the strength of FGSM attacks while maintaining a relatively straightforward implementation -- in comparison to the other methods discussed. However, this approach only alleviates the issue of excessively strong or weak steps. 

Therefore, based on the FGSM, we propose a white-box attack with comparable implementation complexity that relies on \textbf{SH}apley \textbf{A}dditive ex\textbf{P}lanation (SHAP) values as introduced by \textcite{lundberg2016unexpectedunitymethodsinterpreting} and \textcite{lundberg2017unifiedapproachinterpretingmodel} instead of gradients (see equation 5). SHAP values provide information about the extent to which individual inputs contribute to the output in the inference phase. In \cref{sec:shap} we define SHAP values and also discuss the computational intensity to subsequently show in \cref{sec:shap-attacks} why and how SHAP attacks work using four very different data sets and architectures. In \cref{sec:comparison} we compare FGSM from \textcite{kurakin2017adversarialexamplesphysicalworld} with SHAP attacks and in \cref{sec:conclusion} we conclude.

It is crucial to publicly disclose the types of attacks in order to make the models resilient and to establish countermeasures. Without this transparency, attacks could be developed and deployed secretly, potentially harming models currently in use.

\section{SHAP Values}\label{sec:shap}

Post-analytical methods such as \textbf{L}ocal \textbf{I}nterpretable \textbf{M}odel-agnostic \textbf{E}xplanations (LIME) by \textcite{ribeiro2016whyitrustyou} or SHAP values are model-agnostic, meaning they explain the importance of input variables independently from the classifier. SHAP values offer several benefits compared to other post-analytical methods. Unlike LIME, SHAP values can be interpreted both locally and globally. Furthermore, they are additive, meaning that when all individual effects are summed, they precisely match the model's output. 

To ensure the convergence of an additive feature attribution method to a unique solution when explaining model outputs, three properties are essential: 

\textit{Note: Models for the post-analytical explanation of prediction outputs often use simplified inputs \textit{x'} \citep{lundberg2016unexpectedunitymethodsinterpreting}}
\begin{description}
	\item[Local accuracy] ensures that the post-hoc explanation model $g(x')$ for a simplified input $x'$ matches the output of the original model $f(x)$, where $x$ denotes the original input. 
	\item[Missingness] means that a missing input $x$ is treated the same as an input with no effect. 
	\item[Consistency] addresses the issue of multicollinearity. In statistical explanatory models, such as the ordinary least squares method, linear dependence between predictors can be so significant that they change the signs of the coefficients. While this does not impact the prediction accuracy, it affects the interpretation of the model \citep{farrar1967multicollinearity}. Therefore, in a consistent model, if the influence of $x$ increases or remains unchanged, its attributed impact should not decrease.
\end{description}

\textcite{young1985monotonic} demonstrated that Shapley values satisfy the three conditions above. 

SHAP values quantify the impact of a variable \textit{i} on a given task. Starting from the mean of the target variable as the baseline, one model is calculated for each possible subset $S$ of variable combinations for $N$ variables in total. A power set is used to calculate the total contribution $\phi$ of a variable $i$ to the result. This approach involves weighting each individual (marginal) contribution obtained by adding a variable.

For instance, if variables A, B, and C are considered, the process of determining the contribution of A begins at a level without variables (just the average of the target variable) and progresses to a level where a model includes all three variables. Marginal contributions are assessed at each stage, moving from one level to the next. So, the model with only variable A is compared to the baseline (i.e., the average without any variables) and is weighted by one-third. For models with two variables, combinations such as AB and AC are evaluated, each weighted by one-sixth, which sums to one-third in total. The model including all variables is also weighted by one-third. Each marginal contribution of a variable results from the model output including the corresponding variable $f_x(S \lor i)$ and excluding $f_x(S)$.

\begin{equation}
\phi_i = \sum_{S \subseteq N_i} = \frac{(N-|S|-1)!|S|!}{N!}[f_x(S \lor i) - f_x(S)]
\end{equation}

The first factor $\frac{(N-|S|-1)!|S|!}{N!}$ is the weight, the second $[f_x(S \lor i) - f_x(S)]$ is the marginal contribution. This facilitates the determination of how variables contribute to a specific outcome for each observation. However, with $N$ possible variables and $k$\footnote{k is not equalt to S. S contains a certain subset, whereas k is only the number of variables considered at a certain level.} variables included, the number of total possibilities (including the mean of the target) is calculated by
\begin{equation}
\sum_{k=0}^N \binom{N}{k} = 2^N
\end{equation}
This may result in very high computational complexity. Therefore, in practical implementations, a dataset sampling method is employed wherein a constrained least squares problem is addressed using a feasible quantity of data points. However, despite this simplification, high computational costs remain, which is especially notable in the field of computer vision and requires processing with suitable high-performance architecture.

SHAP values are to be understood more as a framework as they use other methods in the background, such as LIME, DeepLift from \textcite{shrikumar2016not}, or layer-wise relevance propagation from \textcite{bach2015pixel}. Eventually, the key strengths of SHAP values stem from their use of the power set, leading to additive contributions, along with a well-defined mathematical foundation.

\begin{figure}[!ht]
\centering
	\includegraphics[width=0.5\textwidth]{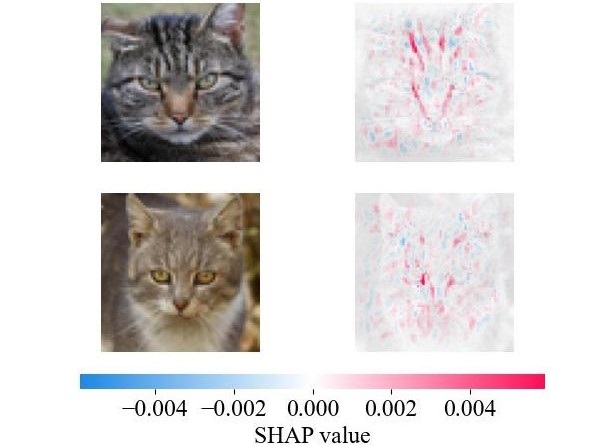}
	\caption{What makes a Cat a Cat?}\label{fig:cat}
\end{figure}

In the case of images, each pixel at channel level (hereinafter referred to as a \enquote{pixel}) can be regarded as a variable $x$. This makes it possible to determine which areas in an image speak for and against the corresponding result in a classification task. \Cref{fig:cat} illustrates the classification process for a dataset consisting of pixels of cats and dogs.

At the bottom, the intensity represents the influence of a specific region: Red shades indicate a higher likelihood of a cat, while the blue shades indicate a lower likelihood of a cat. \Cref{fig:cat} demonstrates that SHAP values can identify which pixels contribute positively or negatively toward a specific class. This indicates that this calculation method could potentially be used to conduct adversarial attacks by modifying the corresponding regions. However, the image does not yet clarify the underlying rules governing this process.

\section{SHAP Attack}\label{sec:shap-attacks}

\begin{figure}[!ht]
	\includegraphics[width=\textwidth]{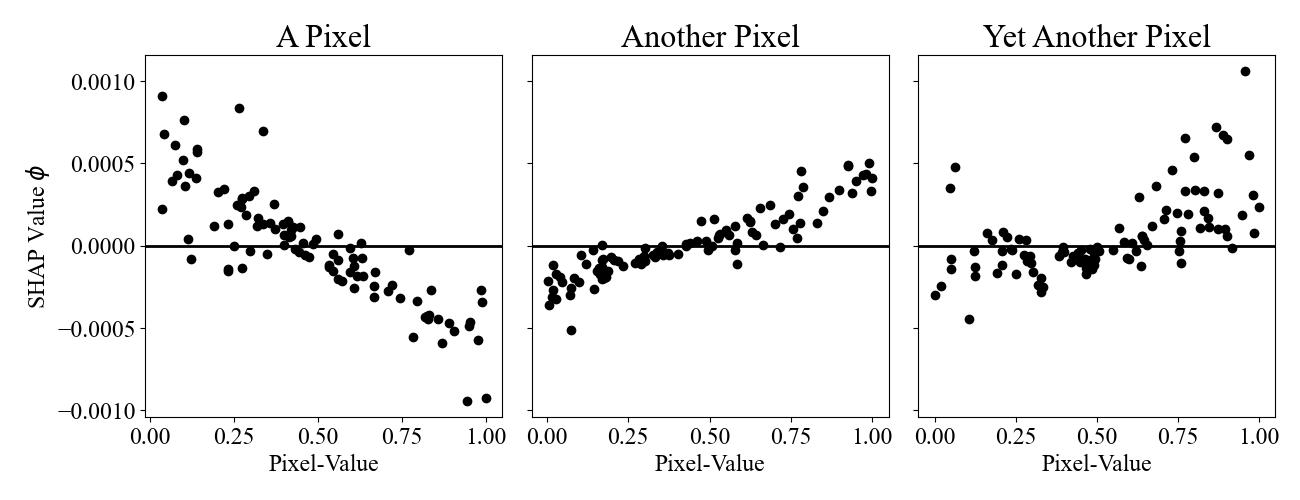}
	\caption{$100$ Images, $3$ different Pixels (SHAP Values vs. underlying Pixel magnitude) for the \enquote{Animal Faces} Dataset.}
	\label{fig:shap-pixel-correlation1}
\end{figure}

Given that SHAP values identify the contributions and magnitudes of individual pixels to the model's output, it is plausible to utilise these values to counteract these influences, potentially inducing misclassifications. The strength over similar attacks, like gradient attacks, is that SHAP values take into account the magnitude of a pixel's influence on the outcome. In contrast, gradient attacks only multiply the sign of the remaining gradient by a scalar $\varepsilon$, ignoring the magnitude of the influence. This means that fewer pixels need to be manipulated, which makes the SHAP attack more subtle.  \Cref{fig:shap-pixel-correlation1} illustrates the relationship between SHAP values and pixel values. 

If a pixel exhibits an exceptionally high or low value, it can exert a considerable influence. In a sufficiently well-fitted classification model, a pixel with an intermediate magnitude tends to have a rather neutral effect. This means that pixels with a value in the middle of the range (between 0-1) are more likely to have a SHAP value close to zero (according to \cref{fig:shap-pixel-correlation1} at a pixel value of approximately 0.50). Analysis of SHAP values across various datasets and architectures listed below confirm this assertion. Additionally, it is important to recognise that a pixel's value is not always positively correlated with its influence on the model output. 

\begin{figure}[!ht]
	\includegraphics[width=\textwidth]{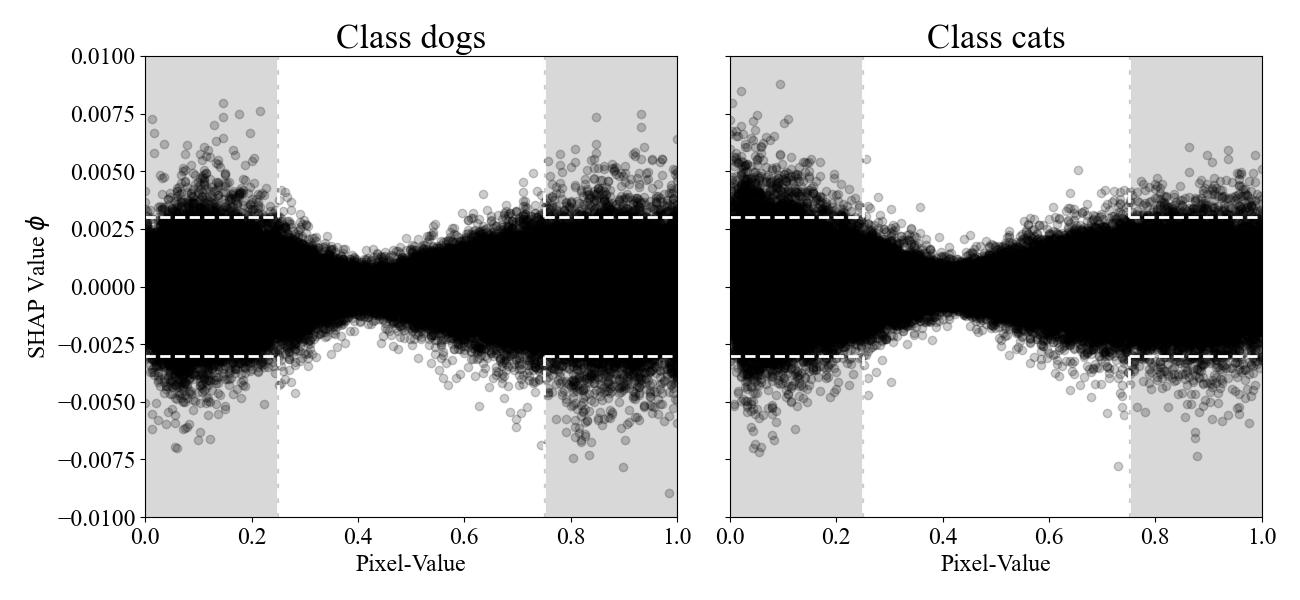}
	\caption{$500$ Images x $64$x$64$x$3$ Pixel (SHAP Values vs. underlying Value) for the \enquote{Animal Faces} Dataset.}
	\label{fig:shap-pixel-correlation2}
\end{figure}

For the magnitude of a single pixel of an image, the relationship to $\phi_i$ often linearly increases or decreases, as depicted in \cref{fig:shap-pixel-correlation1} (A Pixel, Another Pixel) for example. In other occasions the relationship might be convex (Yet Another Pixel \cref{fig:shap-pixel-correlation1}) or concave with either increasing or decreasing tendency. What they all have in common is that they intersect or approach the zero line at approximately the same point. If these relationships, meaning all the pixels from several images, are overlain, it becomes evident that the neutral area mentioned earlier appears somewhere in the centre of the range (in this case 0-1). The result is an almost butterfly pattern, as illustrated in \cref{fig:shap-pixel-correlation2}. In this case, the neutral area is around 0.4.

The grey regions indicate the pixel areas exerting the greatest influence. As discussed in detail below, the butterfly pattern can be leveraged by shifting pixels with positive $\phi$ toward the neutral region, and those with negative $\phi$ toward the corresponding boundaries (0 or 1). This process effectively neutralizes pixels that support a given class while amplifying those that oppose it, thereby increasing classification uncertainty. In order to exclude both architecture and dataset dependencies in the observed pattern, we use different architectures and datasets. The \href{https://www.kaggle.com/datasets/andrewmvd/animal-faces}{\textbf{Animal Faces dataset} from Kaggle} shows portrait photos of cats, dogs and wild animals in 512x512 and three colour channels. We reduce these to 64x64x3. The applied model has 3 convolutional layers and a total of around $76$k parameters. The \href{https://storage.googleapis.com/mledu-datasets/cats_and_dogs_filtered.zip}{the \textbf{Cats and Dogs Filtered dataset} from Google} shows cats and dogs in the format 160x160x3. In contrast to the previous dataset, the dogs and cats are not displayed as portraits, but in different environments and poses, which increases the variance of the dataset in comparison. To examine how SHAP attacks behave with very deep networks, we perform a transfer learning with EfficientNetB7 from \textcite{tan2020efficientnetrethinkingmodelscaling} with more than $66$ million parameters. The \href{https://yann.lecun.com/exdb/mnist/}{\textbf{MNIST}} data set shows handwritten digits from 0-9 in the format 28x28x1 (greyscale). This data set is completely different from all others. The applied classification model has two convolutional layers followed by a dense layer with a total of $450$k parameters.

\begin{figure}[!htb]
	\includegraphics[width=1.02\textwidth]{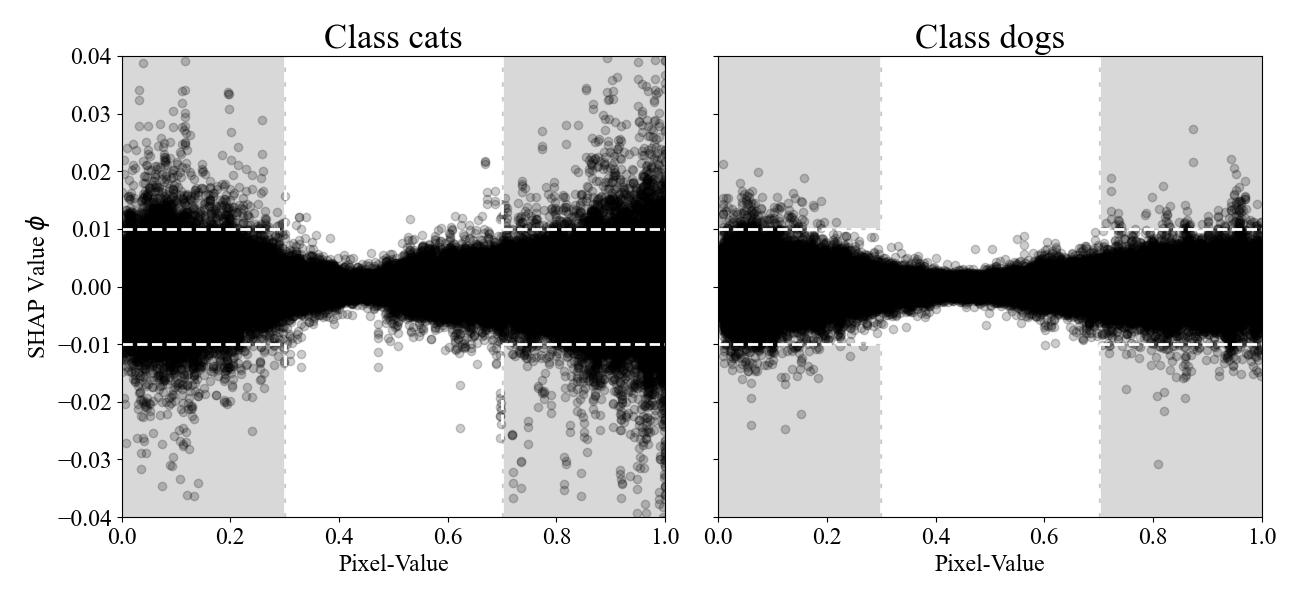}
	\caption{$500$ Images x $160$x$160$x$3$ Pixel (SHAP Values vs. underlying Value) of the \enquote{Cats and Dogs filtered} Dataset.}
	\label{fig:shap-cats-dogs}
\end{figure}

The \href{https://www.kaggle.com/code/frankmollard/efficientnetbx/output?select=s_d.npy}{\textbf{Woman and Man Faces} dataset from Kaggle} contains portraits with male and female appearance in 64x64x3 format. Human faces are very different from animal faces in terms of statistical properties because humans do not have fur to hide skin aging and animals do not wear clothes. Therefore, increased variance is to be expected. The model applied has 3 convolutional layers and a total of around $76$k parameters. 

\Cref{fig:shap-cats-dogs} shows the correlation between SHAP values an pixel values for \href{https://storage.googleapis.com/mledu-datasets/cats_and_dogs_filtered.zip}{the Cats and Dogs Filtered dataset from Google} using the very deep EfficientNetB7.
When applied to the \href{https://yann.lecun.com/exdb/mnist/}{MNIST} data set, the same pattern can be seen for each class.

\begin{figure}[!ht]
	\includegraphics[width=\textwidth]{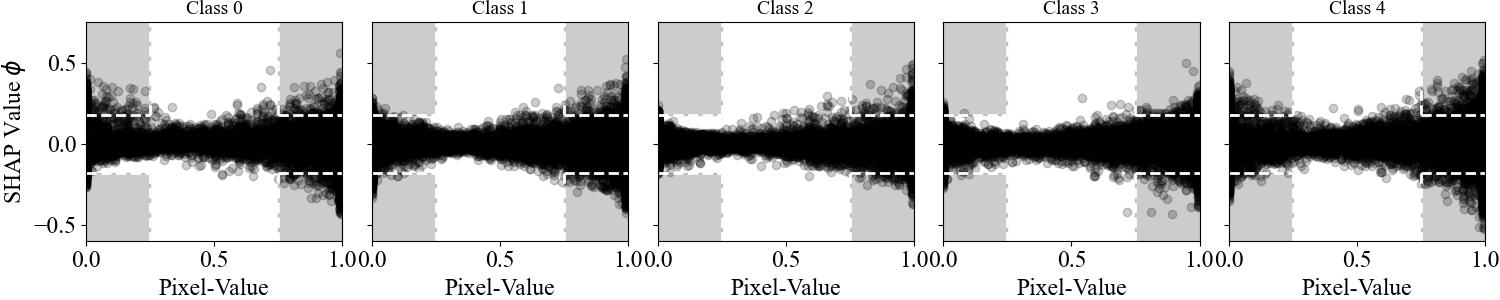}\\ 
	\vskip 0.05cm
	\includegraphics[width=\textwidth]{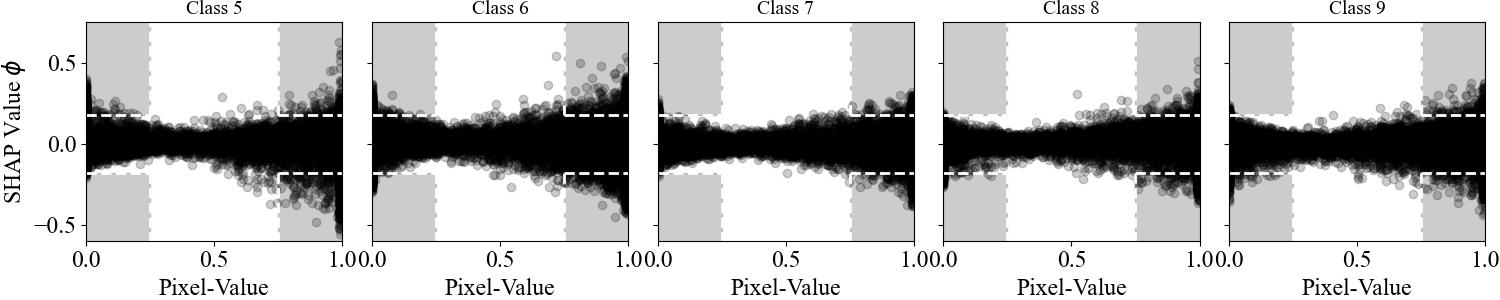}
	\caption{$500$ Images x $28$x$28$ Pixel (SHAP Values vs. underlying Value) \enquote{MNIST} Dataset.}
	\label{fig:MNIST-attack}
\end{figure}

\begin{figure}[!ht]
	\includegraphics[width=\textwidth]{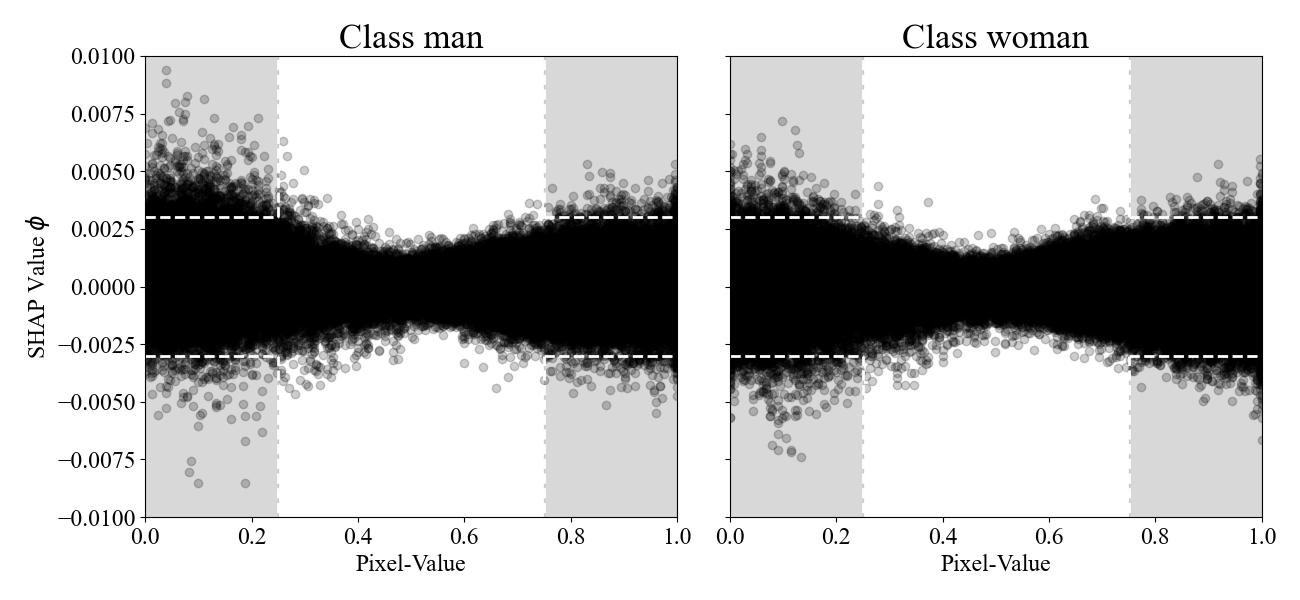}
	\caption{$500$ Images x $160$x$160$x$3$ Pixel (SHAP Values vs. underlying Value) of the \enquote{Woman and Man Faces} Dataset.}
	\label{fig:shap-woman-man}
\end{figure}

It is noticeable in \cref{fig:MNIST-attack} that the area with less influence on the model output is not close to 0.4 as in the two examples above, but rather lower, at around 0.3 or below.
The same pattern can also be observed for human faces, as illustrated in \cref{fig:shap-woman-man}.

Initially, it might seem plausible to neutralise positive influences ($\phi$) on the classification result. However, pixels generally exhibit correlations with one another. If only the positive components were removed, the information would still be available via a simple inverse conclusion. This means that the mere neutralisation of pixel values speaking for a certain class when applying a convolutional neural network does not necessarily lead to misclassification, since the pattern may still persist but only other pixels justify it instead. This is similar to removing all the red areas from \cref{fig:cat}, leaving the same pattern as a systematic gap. It is therefore equally important to strengthen the regions that contradict the class, in order to undermine the model’s confidence. Consequently, moving away from the grey areas for positive SHAP values and shifting the nerative ones to the edges removes the information the model needs for correct classification. In light of this, the SHAP attack can be defined as
\begin{equation}
    V = \phi \geq v \lor \phi \le -v
\end{equation}
\begin{equation} x' = x + \varepsilon \cdot
\begin{cases}
    -|\phi|,& \text{if } x\geq 1-h \:\: \wedge \:\: V \\
    \:\:\: |\phi|,& \text{if } x\le h \:\: \wedge \:\: V \\
\end{cases}
\end{equation}
where $\varepsilon \in [0,\infty]$ stands for the intensity of the attack. In our investigations, $\varepsilon=\frac{1}{20\sigma_{\phi}}$, with $\sigma_{\phi}$ as the standard deviation of $\phi$, proved to be a good starting point. SHAP Values are represented by $\phi$. $v$ is a real numbered vertical threshold to ensure that only strong influences are used for manipulation. We recommend $2\sigma_{\phi}$ for $v$ as a first orientation. The parameter $h$ ensures that only pixels with an increased absolute value are manipulated. Values between 0.2 and 0.3 are recommended.

\section{Comparison between SHAP Attack and FGSM}\label{sec:comparison} 

Adversarial evasion attacks have two requirements. On the one hand, they should lead to misclassification; on the other hand, they should remain invisible to human and machine perception. In the following, a comparison is made between FGSM according to Kurakin et al. (2017) and SHAP attacks. The parameters $\varepsilon$ are adjusted so the attacks remain barely subtle. For this purpose, a visual comparison is made to assess the performance of the two algorithms at the adjusted $\varepsilon$. In addition, a mass evaluation is performed on the data sets to determine the misclassification performance under the circumstances described above.

\subsection{Visual Assessment}

To gauge the degree of subliminality of the different attacks, the \enquote{Animal Faces} dataset is visualised first. The optimal $\varepsilon$, ensuring the attack is imperceptible, is simulated as $0.2$ for FGSM and $50$ for SHAP attacks. 52 \% misclassifications occur with FGSM and 73 \% misclassifications with SHAP attacks. The SHAP attacks have a more sustained effect than FGSM when $\varepsilon$ is increasing.

\begin{figure}[!ht]
\includegraphics[width=\textwidth]{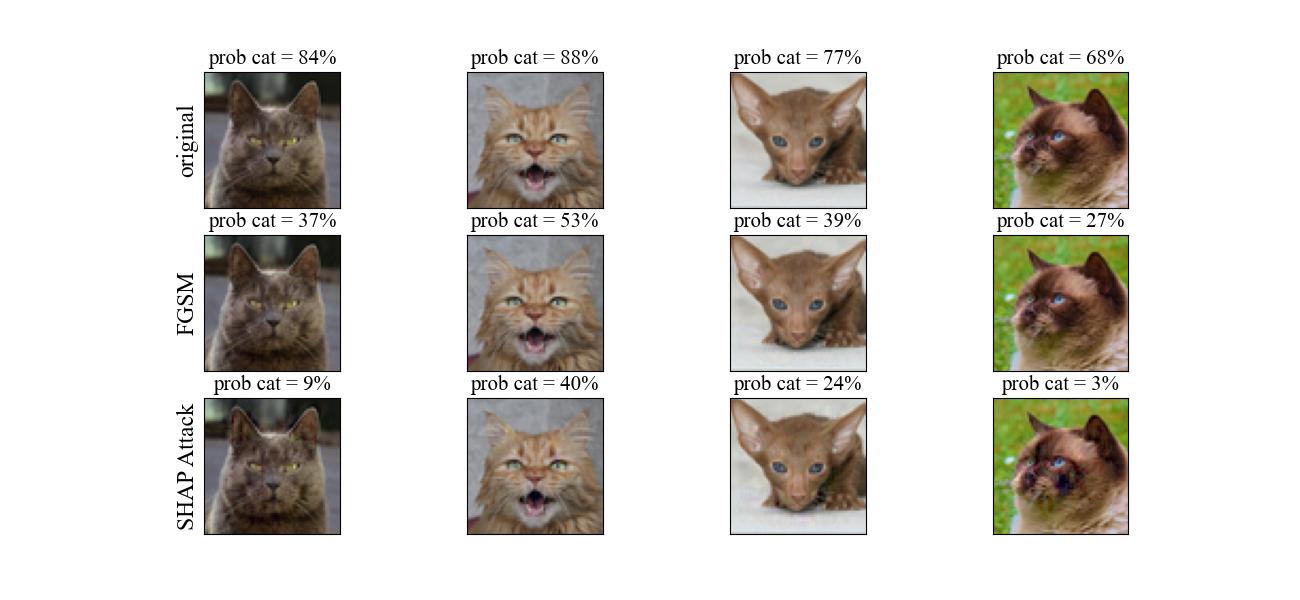}
\caption{$3$ Images x $64$x$64$x$3$ Pixel Attack Comparison on the \enquote{Animal Faces} Dataset.}
\label{fig:fgsm-attack}
\end{figure}

Based on a cut-off of 50 \%, there are fewer misclassifications using FGSM (see e.g. second image in \cref{fig:fgsm-attack}). FGSM does not generally lead to a reduction in the confidence for a specific class. SHAP attacks, however, consistently lead to a reduction in the probability of the attacked class -- assuming that the classification model is sufficiently well-fitted.

\begin{figure}[!ht]
\includegraphics[width=\textwidth]{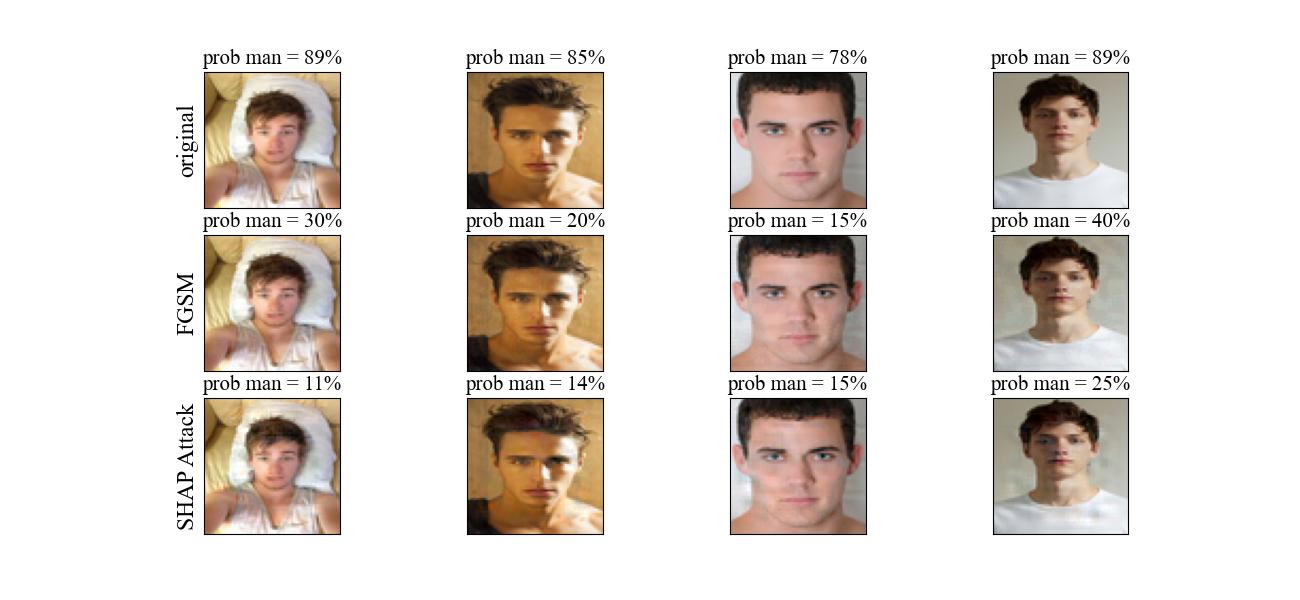}
\caption{$3$ Images x $64$x$64$x$3$ Pixel Attack Comparison on the \enquote{Man and Woman Faces} Dataset.}
\label{fig:shap-attack}
\end{figure}

For human faces \href{https://www.kaggle.com/code/frankmollard/efficientnetbx/output?select=s_d.npy}{(\enquote{Woman and Man Faces} dataset from kaggle)} similar relations are observed. Here $\varepsilon$ is 60 for the SHAP attack and 0.2 for FGSM. The FGSM attack achieved a misclassification rate of 69 \% and SHAP attacks of 98 \%. \Cref{fig:shap-attack} provides a visual demonstration.

Concerning invisibility, it becomes evident (from \cref{fig:fgsm-attack,fig:shap-attack} respectively) that both attacks exhibit comparable effectiveness, though SHAP attacks demonstrate a stron\-ger influence. The stronger misclassification by SHAP attacks in this case is caused by the difference in methodology. While FGSM merely multiplies the gradient sign by a constant, SHAP attacks tend to concentrate on particularly important areas in the image and leave the unimportant areas untouched. Consequently, pixels with significant influence ($\phi$) undergo more pronounced alterations under the SHAP attack methodology.

\subsection{Mass Evaluations}

\Cref{tab:misclassification-rate} presents the rates of misclassification using FGSM by \textcite{kurakin2017adversarialexamplesphysicalworld} and SHAP Attacks. These rates occur at an attack intensity $\varepsilon$, where the manipulation remain imperceptible. With MNIST, imperceptibility is not possible because the manipulation cannot be concealed by the missing channels. Here, the attack was designed in such a way that the perceptibility was slight and comparable between the methods. The classes are balanced in all data sets used for comparison, to ensure that no minority class problem occurs.

\begin{table}[!ht]
\centering
\caption{Misclassification Rate for Evasion Attack using FGSM by \textcite{kurakin2017adversarialexamplesphysicalworld} and SHAP Attack}
\label{tab:misclassification-rate}
\begin{tabular}{rccccc}
\multicolumn{6}{c}{Misclassification Rate}                                                                                                   \\ \hline
\multicolumn{1}{|r|}{Dataset} & \multicolumn{2}{c|}{FGSM}                         & \multicolumn{2}{c|}{SHAP Attack}     &       \multicolumn{1}{c|}{Model Accuracy}              \\ \hline
\multicolumn{1}{|r|}{Animal Faces} & \multicolumn{1}{c|}{52 \%} & \multicolumn{1}{c|}{$\varepsilon=0.2$} & \multicolumn{1}{c|}{73 \%} & \multicolumn{1}{c|}{$\varepsilon=50$} &       \multicolumn{1}{c|}{99.5 \%}    \\ \hline
\multicolumn{1}{|r|}{Cats and Dogs Filtered} & \multicolumn{1}{c|}{98 \%} & \multicolumn{1}{c|}{$\varepsilon=0.01$} & \multicolumn{1}{c|}{89 \%} & \multicolumn{1}{c|}{$\varepsilon=80$} &       \multicolumn{1}{c|}{98 \%} \\ \hline
\multicolumn{1}{|r|}{MNIST} & \multicolumn{1}{c|}{34 \%} & \multicolumn{1}{c|}{$\varepsilon=0.2$} & \multicolumn{1}{c|}{60 \%} & \multicolumn{1}{c|}{$\varepsilon=1.5$} &       \multicolumn{1}{c|}{99 \%}\\ \hline
\multicolumn{1}{|r|}{Woman and Man Faces} & \multicolumn{1}{c|}{69 \%} & \multicolumn{1}{c|}{$\varepsilon=0.2$} & \multicolumn{1}{c|}{98 \%} & \multicolumn{1}{c|}{$\varepsilon=60$} &       \multicolumn{1}{c|}{97 \%}\\ \hline
\end{tabular}
\end{table}

The table illustrates that gradient-based attacks yield variable outcomes, occasionally surpassing expectations but often underperforming. In contrast, SHAP-based attacks consistently result in stable misclassification rates. The instability of gradient-based attacks can be attributed to the gradients becoming uninformative; a phenomenon known as \enquote{gradient masking}. This issue can arise due to stochastic, vanishing, exploding or scattered gradients \citep{gupta2021improvedgradientbasedadversarial}. Furthermore, if the loss function exhibits non-smooth behaviour at the approximated minimum, the residual gradient's direction may be incorrect for vanishing gradients, thus, potentially increasing classification confidence rather than decreasing it. This phenomenon has been shown in the mass evaluations. For some FGSM attacks, the classification confidence has increased instead of decreasing. SHAP attacks are independent of the loss function as they only represent the behaviour of the model-output in relation to certain variables (pixels), as seen in equation 2.

\section{Conclusion}\label{sec:conclusion} 

The study has demonstrated the successful use of SHAP values in conducting white-box adversarial attacks on computer vision models. By utilising the explainability of SHAP values to measure the influence of individual inputs, the proposed SHAP attack method is more robust in causing misclassifications than traditional FGSM attacks. These findings highlight the susceptibility of deep learning models to evasion attacks that subtly alter input data, underscoring the necessity for developing more resilient models to counter such adversarial threats. Additionally, the research reveals the dual potential of SHAP values as tools for both model explanation as well as for the identification and exploitation of model weaknesses. 

Future research should focus on improving the defence mechanisms of computer vision models to mitigate the risks posed by these attacks. A development towards better generalized models would be desirable, where the influence of individual pixels is more balanced across the image and the classification relies less on individual pixels.

However, it should also be noted that SHAP attacks not only require access to the model but also the availability of sufficient inference data for the calculation of the SHAP values. Depending on the resolution of the images and the number of classes, high computing resources may also be required due to the high computational complexity. As a result, the practical application of a large number of very high-resolution images or even videos in combination with complex models may be limited or only manageable with very high computing resources.

\subsection*{Corresponding Author}
\label{sec:cauthor}

Frank Mollard:~\email{f.mollard@aol.com}\\
Business Intelligence \& Data Science, Infraserv GmbH \& Co. Höchst KG, Germany

\subsubsection*{ORCID} 

Frank Mollard\,\orcid{https://orcid.org/0000-0002-6469-4764}\\
Marcus Becker\,\orcid{https://orcid.org/0000-0001-5801-3785}\\
Florian Röhrbein\,\orcid{https://orcid.org/0000-0002-4709-2673}

\printbibliography

\end{document}